\documentclass[sigconf]{acmart}

\usepackage{natbib}
\usepackage{courier}
\usepackage{xcolor}
\usepackage{amsmath}
\usepackage{subcaption}
\usepackage{array,booktabs,makecell}
\usepackage{graphicx}
\usepackage{balance}
\usepackage{comment}
\usepackage [english]{babel}
\usepackage [autostyle, english = american]{csquotes}
\MakeOuterQuote{"}
\setcopyright{acmcopyright}
\copyrightyear{2018}
\acmYear{2018}
\acmDOI{XXXXXXX.XXXXXXX}

\acmConference[NLPIR 2022]{NLP }{December}{Bangkok, Thailand}


\copyrightyear{2022}
\acmYear{2022}
\setcopyright{acmlicensed}\acmConference[NLPIR 2022]{2022 6th International Conference on Natural Language Processing and Information Retrieval}{December 16--18, 2022}{Bangkok, Thailand}
\acmBooktitle{2022 6th International Conference on Natural Language Processing and Information Retrieval (NLPIR 2022), December 16--18, 2022, Bangkok, Thailand}
\acmPrice{15.00}
\acmDOI{10.1145/3582768.3582777}
\acmISBN{978-1-4503-9762-9/22/12}

\begin{document}

\title{Explaining Math Word Problem Solvers}

\author{Abby Newcomb}
\email{abbynewcomb13@gmail.com}
\affiliation{%
  \institution{St. Olaf College}
  \streetaddress{1500 St. Olaf Ave}
  \city{Northfield}
  \state{Minnesota}
  \country{USA}
  \postcode{550571}
}

\author{Jugal Kalita}
\email{ljkalita@uccs.edu}
\affiliation{%
  \institution{University of Colorado, Colorado Springs}
  \streetaddress{1420 Austin Bluffs Pkwy}
  \city{Colorado Springs, CO}
  \country{USA}
}

\renewcommand{\shortauthors}{Newcomb and Kalita}


\begin{abstract}
Automated math word problem solvers based on neural networks have successfully managed to obtain 70-80\% accuracy in solving arithmetic word problems. However, it has been shown that these solvers may rely on superficial patterns to obtain their equations. In order to determine what information math word problem solvers use to generate solutions, we remove parts of the input and measure the model's performance on the perturbed dataset. Our results show that the model is not sensitive to the removal of many words from the input and can still manage to find a correct answer when given a nonsense question. This indicates that automatic solvers do not follow the semantic logic of math word problems, and may be overfitting to the presence of specific words.
\end{abstract}

\maketitle


\section{Introduction}

Math word problem (MWP) solving is an area of natural language processing (NLP) that uses machine learning to solve simple arithmetic problems. MWPs consist of a few sentences of text including a few numbers and an unknown quantity, similar to problems humans are presented with in grade school. Neural networks are trained to generate the correct equation which computes the unknown quantity. Little is known about \textit{how} neural networks manage to solve math word problems. In this paper, we remove parts of math word problems and measure the model's performance on the changed data in order to ascertain which words the model is using to choose the correct equation.

Various parts of speech work together to construct the full meaning of a sentence, so even when a certain part of speech is removed, other words may still indicate the desired operation. In order to more specifically gauge which words are important to the model's prediction, we employ input reduction, a strategy that iteratively removes the least important word from the input until the model produces an incorrect result. This method allows us to see how removing specific words affects the model.

We also perform analysis of which words appear most frequently in the datasets used to train the model. We also look at the most common words for each type of problem (+, -, *, /, multiple) to see whether certain words appear to indicate specific operations.

In order to determine which parts of speech are most important to MWP solvers, we remove specific words from MWP test datasets and test a Seq2seq MWP solver on its ability to determine the correct answer on these reduced problems. The contributions of this paper are as follows:
\begin{itemize}
    \item We show that the lexical diversity of MaWPS is low.
    \item We show that the RNN Seq2seq solver performs little semantic reasoning, since it can produce correct answers with significantly reduced input.
\end{itemize}

We begin by explaining related work, then cover the methods and results of each experiment in turn, followed by the conclusion.

\begin{figure}
\centering
  \includegraphics[scale=.65]{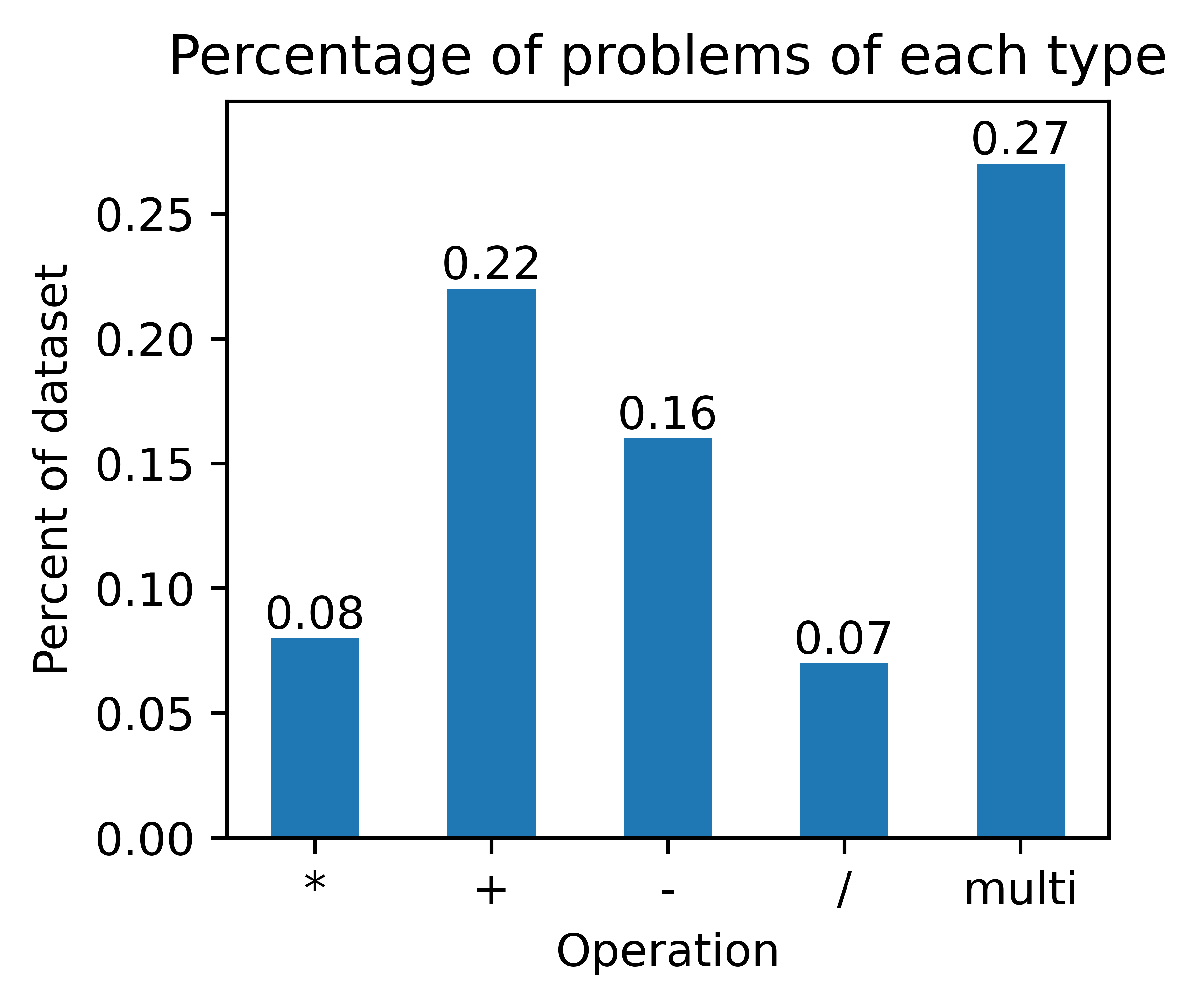}
  \caption{Percentage of MWPs in MaWPS dataset of each operation type, on average across all CV folds.}
  \label{fig:mawps_op_pct}
\end{figure}

\section{Related Work}

Various neural network MWP solvers have been created and benchmarked on well-known datasets. Few explainability techniques have yet been applied to MWP solving.

\subsection{Math Word Problems}

The current most commonly used datasets for Math Word Problem solving are MaWPS \cite{koncel-kedziorski_mawps_2016} and ASDiv-A \cite{miao_diverse_2021}. These datasets are currently the largest ones available, though they are quite small for machine learning datasets. MaWPS has 2373 MWPs while ASDiv-A has only 1218 problems.

\begin{figure*}
\centering
  \includegraphics[scale=.65]{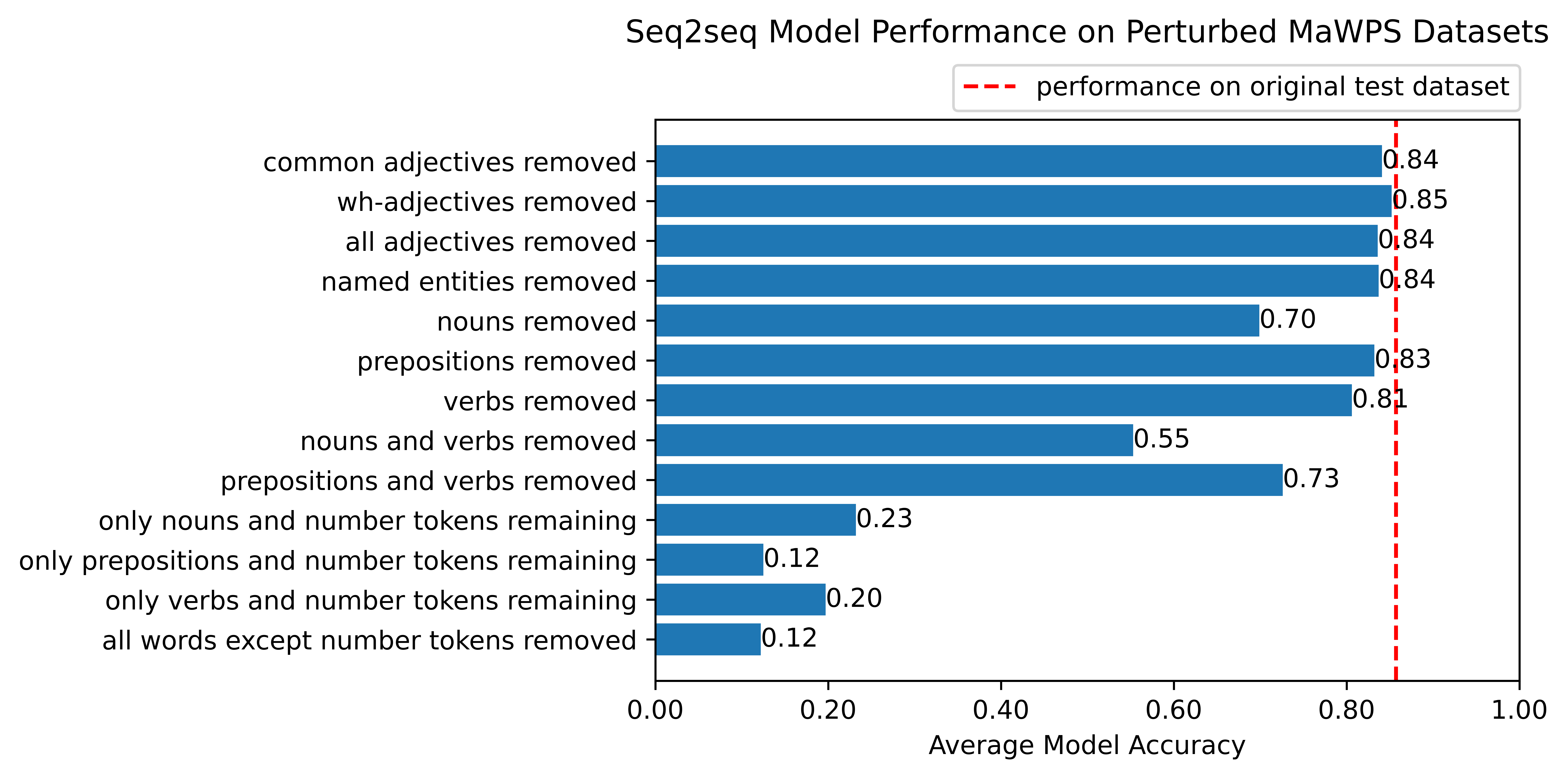}
  \caption{The average accuracy of the Seq2seq model trained on the MaWPS dataset, when evaluated on various perturbed datasets. The model's average accuracy on the original test dataset is indicated by the red dashed line.}
  \label{fig:rnn_seq2seq_performance_mawps}
\end{figure*}

Various types of neural networks for solving MWPs have been developed.
\citeauthor{wang_deep_2017} (\citeyear{wang_deep_2017}) use a GRU encoder and an LSTM decoder in a sequence to sequence approach.  Another model is a graph to tree model proposed by \citeauthor{zhang_graph--tree_2020} (\citeyear{zhang_graph--tree_2020}), which uses a graph transformer and tree structured decoder to generate the MWP solution expression tree. \citeauthor{griffith_solving_2019} (\citeyear{griffith_solving_2019}) use a transformer-based model. \citeauthor{xie_goal-driven_2019} (\citeyear{xie_goal-driven_2019}) use a model called GTS in a process they call goal decomposition to find relationships between quantities. Their approach uses feed-forward networks and an RNN model at different steps in the algorithm. 

Though these models obtain high accuracy, their success was called into question when MWP solvers were shown to obtain similar accuracy when the actual question was removed, leaving only the descriptive body of text at the beginning of the problem \cite{patel_are_2021}. MWP solvers also perform poorly on the SVAMP challenge dataset, which was specifically generated to require attention to the question itself \cite{patel_are_2021}. This implies that the solvers are relying on superficial patterns in the initial text rather than actually answering the question posed in the problem. However, it was later shown that performance on the SVAMP dataset could be improved simply by generating more data to increase the size of MWP training datasets \cite{kumar_practice_2022}.

\begin{figure*}
\centering
  \includegraphics[scale=.65]{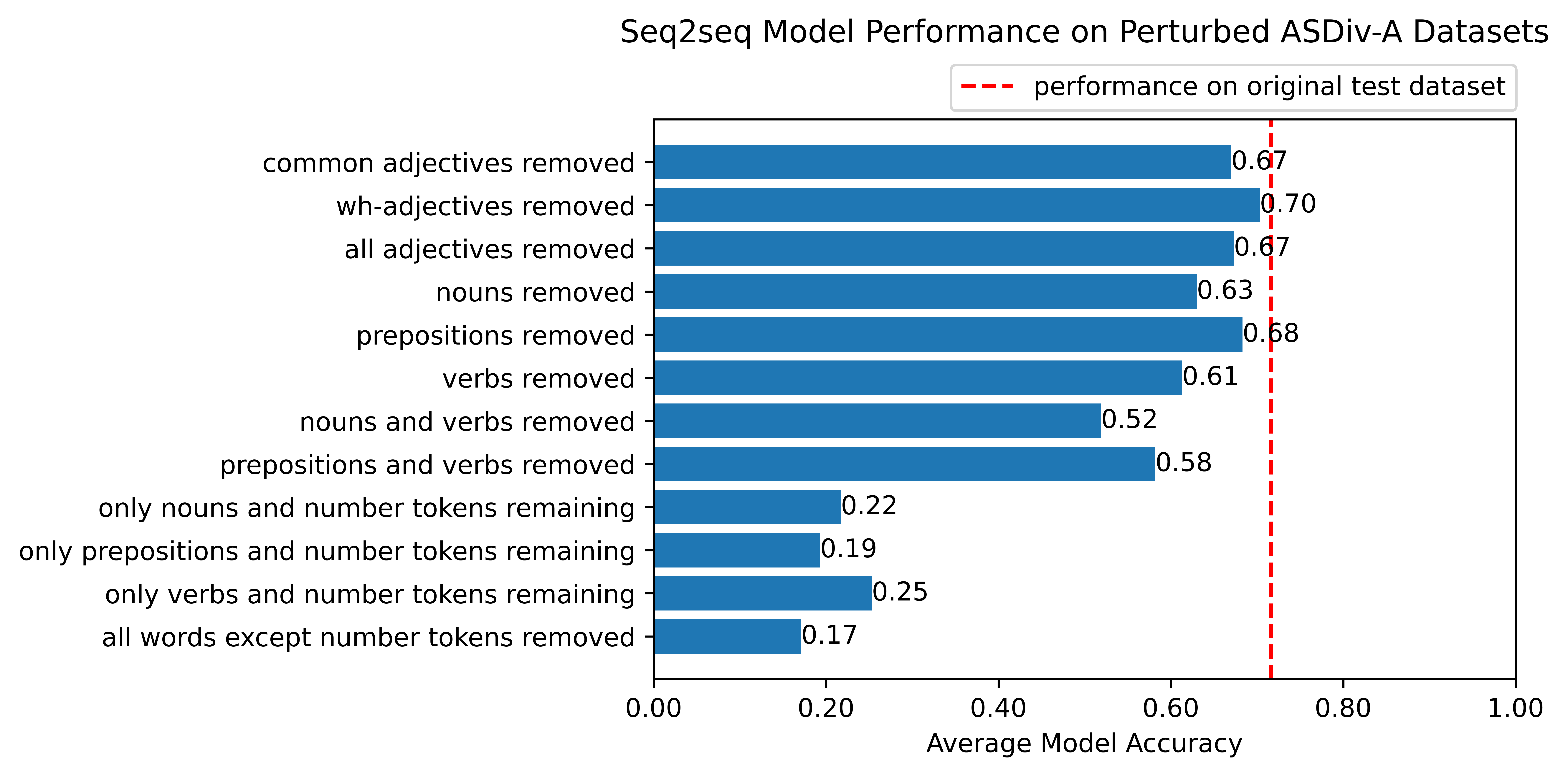}
  \caption{The average accuracy of the Seq2seq model trained on the ASDiv-A dataset, when evaluated on various perturbed datasets. The model's average accuracy on the original test dataset is indicated by the red dashed line.}
  \label{fig:rnn_seq2seq_performance_asdiv}
\end{figure*}
\subsection{Explainability Techniques}

The strategy of removing parts of the input to an NLP model is often used to explain a model's decisions. Importance scores have been assigned to words in the input by looking at the effects of removing those words \cite{li_understanding_2017}. Similarly, the process of input reduction involves successively removing the word that affects the model's confidence score the least, until we are left with the smallest possible input with which the model can still make a correct prediction \cite{feng_pathologies_2018}. This process shows us which words in the input are most important to the model's prediction. These methods, among others, have been implemented by \citeauthor{wallace_allennlp_2019} (\citeyear{wallace_allennlp_2019}) in their AllenNLP framework for NLP explainability techniques. However, applications of these methods often focus on large models such as BERT and tasks such as to sentiment analysis, reading comprehension, or textual entailment. This method has not yet been applied to MWP solving.

Another method of understanding NLP model predictions is adversarial attacks, in which various changes are made to the input of a model, and the performance of the model is measured in order to determine how sensitive the model is to the changes in the perturbed dataset. Adversarial attacks are different from the aforementioned methods because the new inputs to the model are meant to be semantically equivalent to the previous inputs and should still be grammatically correct \cite{lee_qadiver_2019}. Adversarial examples have been used for interpretability of reading comprehension systems \cite{jia_adversarial_2017} and question answering systems \cite{lee_qadiver_2019} in the past. Adversarial attacks involving question reordering and sentence paraphrasing were also used by
\citeauthor{kumar_adversarial_2021} (\citeyear{kumar_adversarial_2021}) to show that MWP solvers are not robust to these seemingly irrelevant perturbations.

\begin{table*}
    \centering
    \begin{tabular}{ccccc}
        \begin{subtable}[h]{0.167\textwidth}
        \begin{tabular}{|p{.21in}rr|}
        \midrule
        \midrule
               &   Count &   Pct \\
        \midrule
         book  &      68 &  0.16 \\
         will  &      62 &  0.14 \\
         were  &      61 &  0.14 \\
         box   &      52 &  0.12 \\
         tree  &      52 &  0.12 \\
         total &      51 &  0.12 \\
         at    &      49 &  0.11 \\
         is    &      49 &  0.11 \\
         pick  &      48 &  0.11 \\
         from  &      46 &  0.11 \\
         park  &      45 &  0.1  \\
        \midrule
        \midrule
        \end{tabular}
        \caption{Addition}
        \label{tab:add}
        \end{subtable} &
        \begin{subtable}[h]{0.167\textwidth}
        \begin{tabular}{|p{.21in}rr|}
        \midrule
        \midrule
                 &   Count &   Pct \\
        \midrule
         dollar  &      63 &  0.19 \\
         total   &      52 &  0.16 \\
         game    &      44 &  0.13 \\
         balloon &      43 &  0.13 \\
         book    &      41 &  0.12 \\
         will    &      40 &  0.12 \\
         at      &      39 &  0.12 \\
         were    &      39 &  0.12 \\
         pick    &      37 &  0.11 \\
         is      &      37 &  0.11 \\
         all     &      36 &  0.11 \\
        \midrule
        \midrule
        \end{tabular}
        \caption{Subtraction}
        \label{tab:sub}
        \end{subtable} &
        \begin{subtable}[h]{0.167\textwidth}
        \begin{tabular}{|p{.21in}rr|}
        \midrule
        \midrule
               &   Count &   Pct \\
        \midrule
         card  &      30 &  0.18 \\
         were  &      26 &  0.16 \\
         box   &      25 &  0.15 \\
         will  &      25 &  0.15 \\
         now   &      23 &  0.14 \\
         total &      23 &  0.14 \\
         book  &      22 &  0.13 \\
         from  &      22 &  0.13 \\
         pick  &      21 &  0.13 \\
         one   &      21 &  0.13 \\
         all   &      20 &  0.12 \\
        \midrule
        \midrule
        \end{tabular}
        \caption{Multiplication}
        \label{tab:mult}
        \end{subtable} &
        \begin{subtable}[h]{0.167\textwidth}
        \begin{tabular}{|p{.21in}rr|}
        \midrule
        \midrule
                 &   Count &   Pct \\
        \midrule
         piece    &      23 &  0.16 \\
         his      &      23 &  0.16 \\
         dollar  &      21 &  0.15 \\
         box     &      21 &  0.15 \\
         from    &      19 &  0.14 \\
         make    &      18 &  0.13 \\
         hour    &      18 &  0.13 \\
         at      &      17 &  0.12 \\
         now     &      16 &  0.11 \\
         balloon &      16 &  0.11 \\
         game    &      15 &  0.11 \\
        \midrule
        \midrule
        \end{tabular}
        \caption{Division}
        \label{tab:div}
        \end{subtable} &
        \begin{subtable}[h]{0.167\textwidth}
        \begin{tabular}{|p{.21in}rr|}
        \midrule
        \midrule
                &   Count &   Pct \\
        \midrule
         dollar &      89 &  0.16 \\
         box    &      77 &  0.14 \\
         piece   &      73 &  0.13 \\
         book   &      69 &  0.13 \\
         total  &      69 &  0.13 \\
         at     &      65 &  0.12 \\
         will   &      63 &  0.11 \\
         all    &      61 &  0.11 \\
         game   &      61 &  0.11 \\
         from   &      61 &  0.11 \\
         would  &      61 &  0.11 \\
        \midrule
        \midrule
        \end{tabular}
        \caption{Multiple}
        \label{tab:multi}
        \end{subtable} \\
    \end{tabular}
    \caption{The top words for each operation in MaWPS CV Fold 1, excluding words that appeared in all 5 lists, by count of MWPs it appears in. Percentage of MWPs of that operation that the word appears in is also provided for comparison's sake.}
    \label{tab:top_words}
\end{table*}

\section{Problem Statement}

The question remains to what degree MWP solvers perform semantic reasoning, and what information they use to generate an equation for a solution to a given problem. We apply various methods to search for trigger words and other superficial patterns that the model may be relying on instead of semantic reasoning.

\section{Experiment 1: Removing Parts of Speech}

We removed various parts of speech from the MWPs and tested an MWP solver's performance on the perturbed datasets in order to see how important different types of words are to the model. A large decrease in accuracy due to the removal of a part of speech indicates that that part of speech is important to the model's prediction, since the model cannot perform as well without it.

\subsection{Methods}

We generate perturbed MWPs by identifying parts of speech using the Natural Language Toolkit (NLTK) part-of-speech tagger \cite{bird_natural_2009} and then removing the targeted words. We use the Seq2seq model created by \citeauthor{patel_are_2021} (\citeyear{patel_are_2021}) for all experiments. We also use \citeauthor{patel_are_2021} (\citeyear{patel_are_2021})'s optimized parameters for training. Two models were trained on either MaWPS or AsDIV-A with 5 fold cross-validation, and then each was evaluated on perturbed examples from its respective dataset. Accuracy is measured on the model's success in generating the correct answer, rather than by the proximity of the generated equation to the true equation.

\begin{figure}
\centering
  \includegraphics[scale=.55]{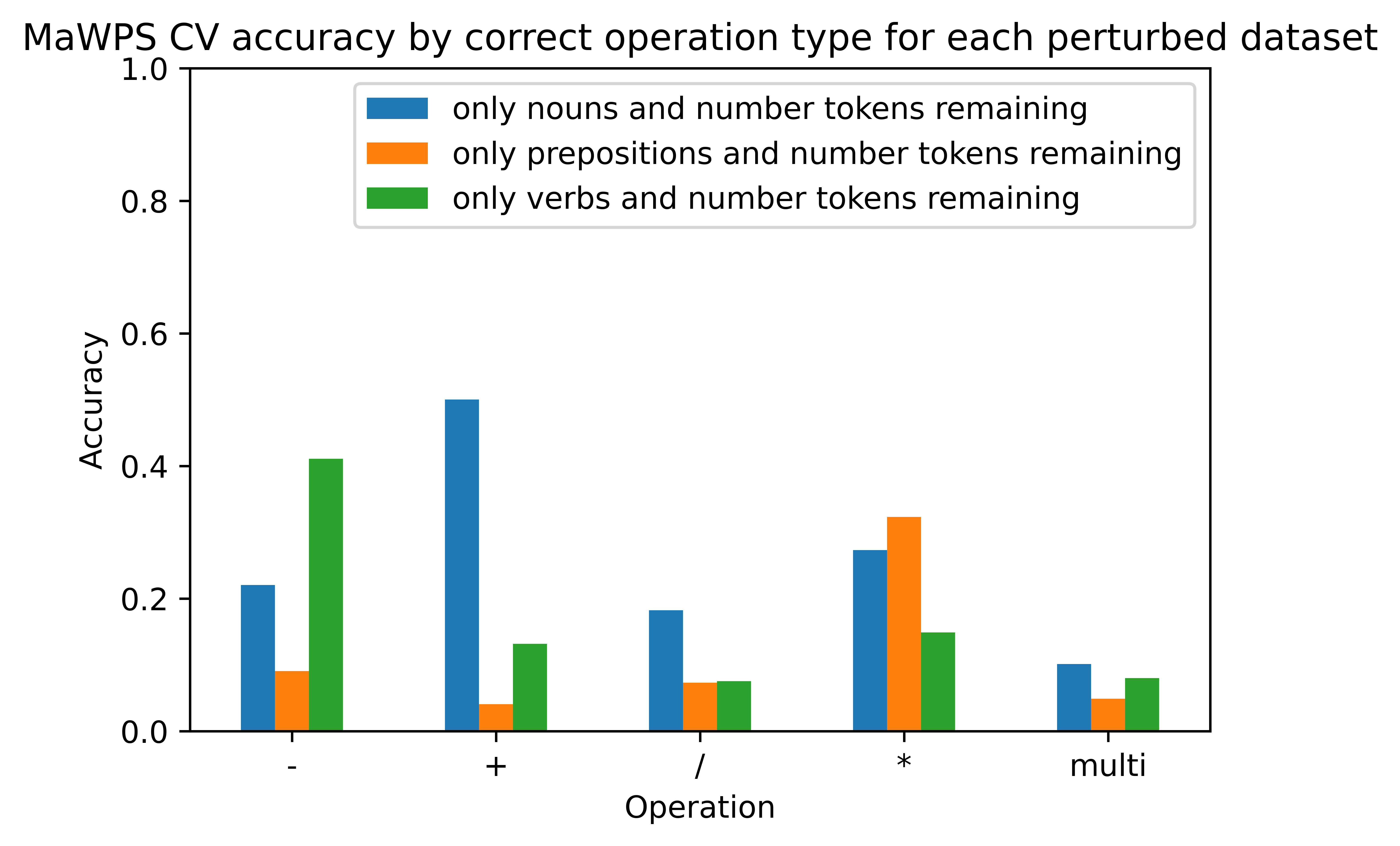}
  \caption{The RNN Seq2seq model's accuracy on each type of problem for the perturbed datasets with only one part of speech and number tokens remaining.}
  \label{fig:mawps_accuracy_by_op_add}
\end{figure}

As a bit of preliminary analysis, we looked at the relative concentration of different types of MWPs in MaWPS, as seen in Figure \ref{fig:mawps_op_pct}. The first four categories are characterized by having a single operation of the specified type, while the problems in the "multi" category have multiple operations of different types in them. The majority of problems in MaWPS (73\%) have only one operation. The dataset appears to represent addition and subtraction the best, and have a much smaller number of multiplication and division problems. This may contribute to the slightly decreased accuracy on multiplication and division problems visible on Figure \ref{fig:mawps_accuracy_by_op_add}.

\subsection{Results}

\begin{table*}
    \centering
    \begin{tabular}{p{2.5cm}@{\hskip .55cm} | p{5cm} | p{4cm} | p{3cm}}
        \midrule\midrule
        \textbf{Perturbation}  
        & \textbf{Original Question}       
        & \textbf{Perturbed Question}  
        & \textbf{Correct Equation}   
        \\\midrule
        Verbs Removed
        & Tommy had some balloons . His mom gave him number0 more balloons for his birthday . Then , Tommy had number1 balloons . How many balloons did Tommy have to start with ?	
        & Tommy some balloons . His mom him number0 more balloons for his birthday . Then , Tommy number1 balloons . How many balloons Tommy to with ?
        & - number1 number0
        \\\midrule
        Nouns Removed
        &The first minute of a telephone call costs number0 cents and each additional minute number1 cents . What is the cost of a number2 minute telephone call ?	
        & The first of a number0 and each additional number1 . What is the of a number2 ?
        & + number0 * number1 number2
        \\\midrule
        Nouns and Verbs Removed        
        & Virginia starts with number0 eggs . Amy takes number1 away . How many eggs does Virginia end with ? 
        & with number0 . number1 away . How many with ? 
        & - number0 number1  
        \\\midrule
        Prepositions and Verbs Removed 
        & In March it rained number0 inches . It rained number1 inches less in April than in March . How much did it rain in April ? 
        & March it number0 inches . It number1 inches less April March . How much it April ?  
        & - number0 number1 
        \\\midrule\midrule
    \end{tabular}
    \caption{Examples of perturbed MWPs from the MaWPS dataset. In this dataset, the actual numbers are removed and replaced with number tokens ("number0", "number1", etc.) in order for the model to process them more easily.}
    \label{tab:example_p_qs}
\end{table*}

Some examples of perturbed MWPs obtained from the MaWPS dataset are shown in   Table  \ref{tab:example_p_qs}.
The models' accuracy on each perturbed dataset is listed in Figures \ref{fig:rnn_seq2seq_performance_mawps} and \ref{fig:rnn_seq2seq_performance_asdiv}, and Table \ref{tab:p_res} shows all percent accuracies and decreases in accuracy. On the original dataset, the model trained on ASDiv-A had 72.4\% accuracy while the MaWPS model had 86.5\% accuracy. Removal of common adjectives such as "more" resulted in accuracy decreases of 5.4\% and 2.4\% respectively, while removing question adjectives such as "how" decreased accuracy by only 2.1\% and 1.3\%, and removal of all adjectives decreased accuracy by 5.1\% and 2.9\%. Removal of named entities such as "Jim" was only conducted with MaWPS because of the formatting of the data. MaWPS model accuracy decreased by only 2.8\% with no named entities. Removal of all nouns, including named entities and all common nouns, decreased accuracy by 9.4\% on ASDiv-A and 16.6\% on MaWPS. Removing prepositions decreased accuracy by 4.1\% and 3.3\% respectively. Removing verbs decreased accuracy by 11.1\% in the ASDiv-A model and by 5.9\% on the MaWPS model. 

\begin{table*}
    \centering
    \begin{tabular}{l| p{1.8cm} p{1.8cm}|p{1.8cm}p{1.8cm}}
    \midrule\midrule
       Perturbation                     &  MaWPS CV Accuracy  &   MaWPS Decrease in Accuracy &   ASDiv-A CV Accuracy &   ASDiv-A Decrease in Accuracy \\
    \midrule
     original dataset                              &           0.857 &            -     &             0.716 &              -     \\
     common adjectives removed                     &           0.841 &            0.017 &             0.67  &              0.046 \\
     wh-adjectives removed                         &           0.852 &            0.004 &             0.703 &              0.013 \\
     all adjectives removed                        &           0.836 &            0.021 &             0.673 &              0.043 \\
     named entities removed                        &           0.837 &            0.02  &           -     &            -     \\
     nouns removed                                 &           0.699 &            0.158 &             0.63  &              0.086 \\
     prepositions removed                          &           0.832 &            0.025 &             0.683 &              0.033 \\
     verbs removed                                 &           0.806 &            0.051 &             0.613 &              0.103 \\
     nouns and verbs removed                       &           0.553 &            0.304 &             0.519 &              0.197 \\
     prepositions and verbs removed                &           0.726 &            0.13  &             0.582 &              0.134 \\
     only nouns and number tokens remaining        &           0.232 &            0.625 &             0.217 &              0.499 \\
     only prepositions and number tokens remaining &           0.125 &            0.732 &             0.193 &              0.523 \\
     only verbs and number tokens remaining        &           0.197 &            0.66  &             0.253 &              0.463 \\
     all words except number tokens removed        &           0.122 &            0.735 &             0.171 &              0.545 \\
    \midrule\midrule
    \end{tabular}
    \caption{Seq2seq model CV accuracy and decrease in CV accuracy on each perturbed dataset.}
    \label{tab:p_res}
\end{table*}

We also tested the models on datasets with two different parts of speech missing. On a dataset with all nouns and verbs missing, the ASDiv-A model accuracy decreased by 20.5\% and MaWPS by 31.2\%. With all prepositions and verbs removed, the models' accuracy decreased by 14.2\% and 13.9\% respectively.

The model was also tested on datasets where only a specific part of speech and the number tokens were left in the MWP, with all other words removed from the input. The results on these datasets tended to somewhat mirror the model's performance on the datasets with that part of speech removed. 

On a dataset with all words except for the number tokens removed, the model achieved 12.2\% accuracy for MaWPS and 17.1\% accuracy on AsDIV-A. It is difficult to calculate what a completely random accuracy would be and how close these are to random guesses because of the complexity of multiple operations, but the AsDIV-A model does manage a significantly higher accuracy, which indicates that it may not rely on the word content as much as the MaWPS model.

\subsection{Discussion}

The model's overall higher accuracies on the MaWPS dataset can likely be attributed to its size, since with 2373 MWPs it is nearly twice as large as ASDiv-A's 1218 problems. The MaWPS model was also less affected by the removal of any single part of speech compared to the ASDiv-A model (average accuracy difference of 5.0\% to ASDiv-A's 6.2\%), and thus seems to be less sensitive to this type of perturbation overall. The MaWPS model was also more affected by the removal of multiple parts of speech, as the decrease in performance on the twice perturbed datasets was larger than the sum of the decrease in performance on either of the once perturbed datasets, which was not the case for the ASDiv-A model. % terribly explained but no time to clarify / I don't feel like expending the mental energy to explain concisely

\begin{figure*}
\centering
  \includegraphics[scale=.64]{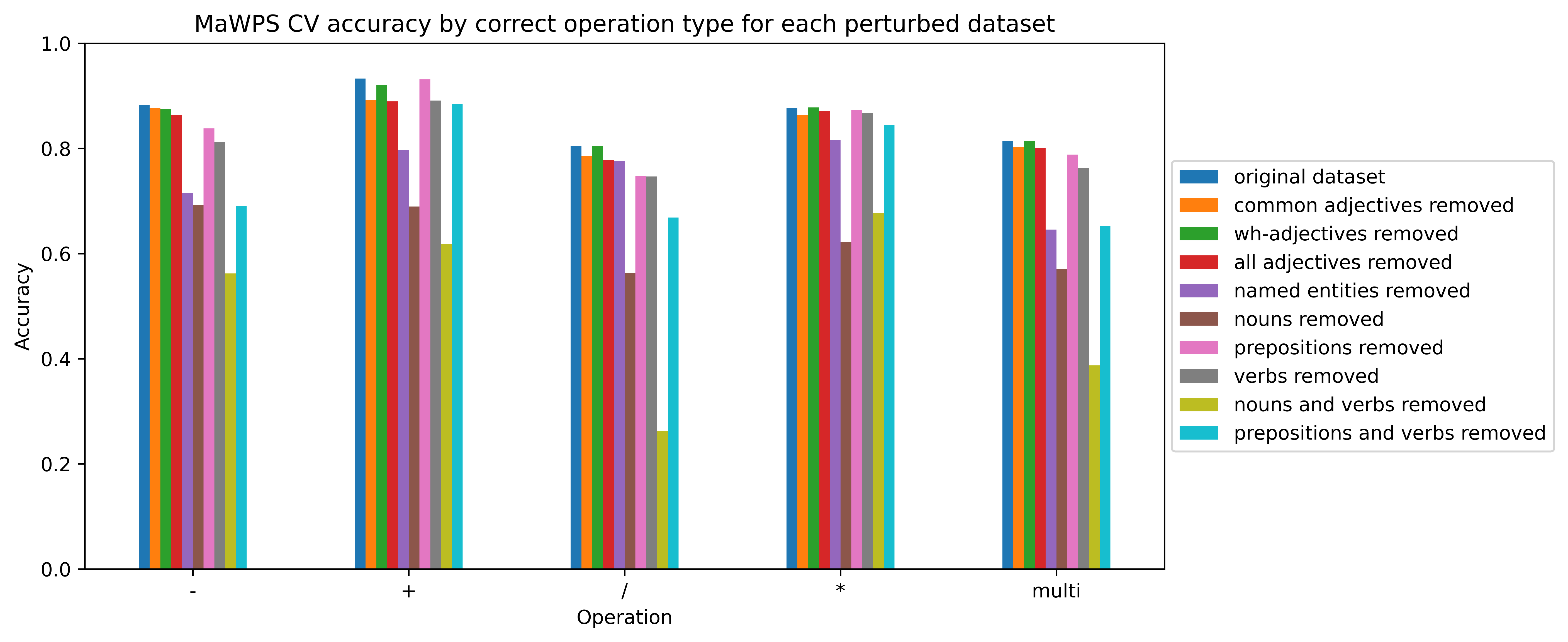}
  \caption{The RNN Seq2seq model's accuracy on each type of problem for the perturbed datasets with one or two parts of speech missing.}
  \label{fig:mawps_accuracy_by_op_rmv}
\end{figure*}

The removal of any single part of speech does not appear to significantly affect either model. Overall, the MaWPS model was most affected by the removal of nouns at a 16.6\% decrease in accuracy, and the ASDiv-A model was most affected by the removal of verbs at an 11.1\% decrease. As hypothesized, certain operations are more affected by the removal of some parts of speech more than others, as seen in Figures \ref{fig:mawps_accuracy_by_op_add} and \ref{fig:mawps_accuracy_by_op_rmv}. The models' decent performance on these reduced datasets indicates that no single part of speech is incredibly important to its decision.

However, both models were still achieving an accuracy above 50\% with no nouns or verbs in the MWPs. This relatively high accuracy indicates that these models are likely not performing semantic reasoning about the events described in the MWP, since there is not enough information in the problem with no verbs or nouns for the model to truly be reasoning about the quantities present. Instead, the solver may be relying on the presence of trigger words. For example, the words "more" and "together" are likely to signal addition even if the model is given no additional context, while "each" may signal multiplication or division. 

For the datasets with only one part of speech and the number tokens remaining, no extremely large jumps in accuracy were observed that would suggest that the model relies entirely on one part of speech. However, accuracy was nearly doubled from 12\% to 23\% with only nouns in the MaWPS model, which does suggest at least some reliance on the presence of certain nouns in this model since clearly with only nouns to go draw its conclusions, no logical reasoning of events is possible.

\section{Experiment 2: MaWPS Word Frequency}

\begin{figure}
\centering
  \includegraphics[scale=.55]{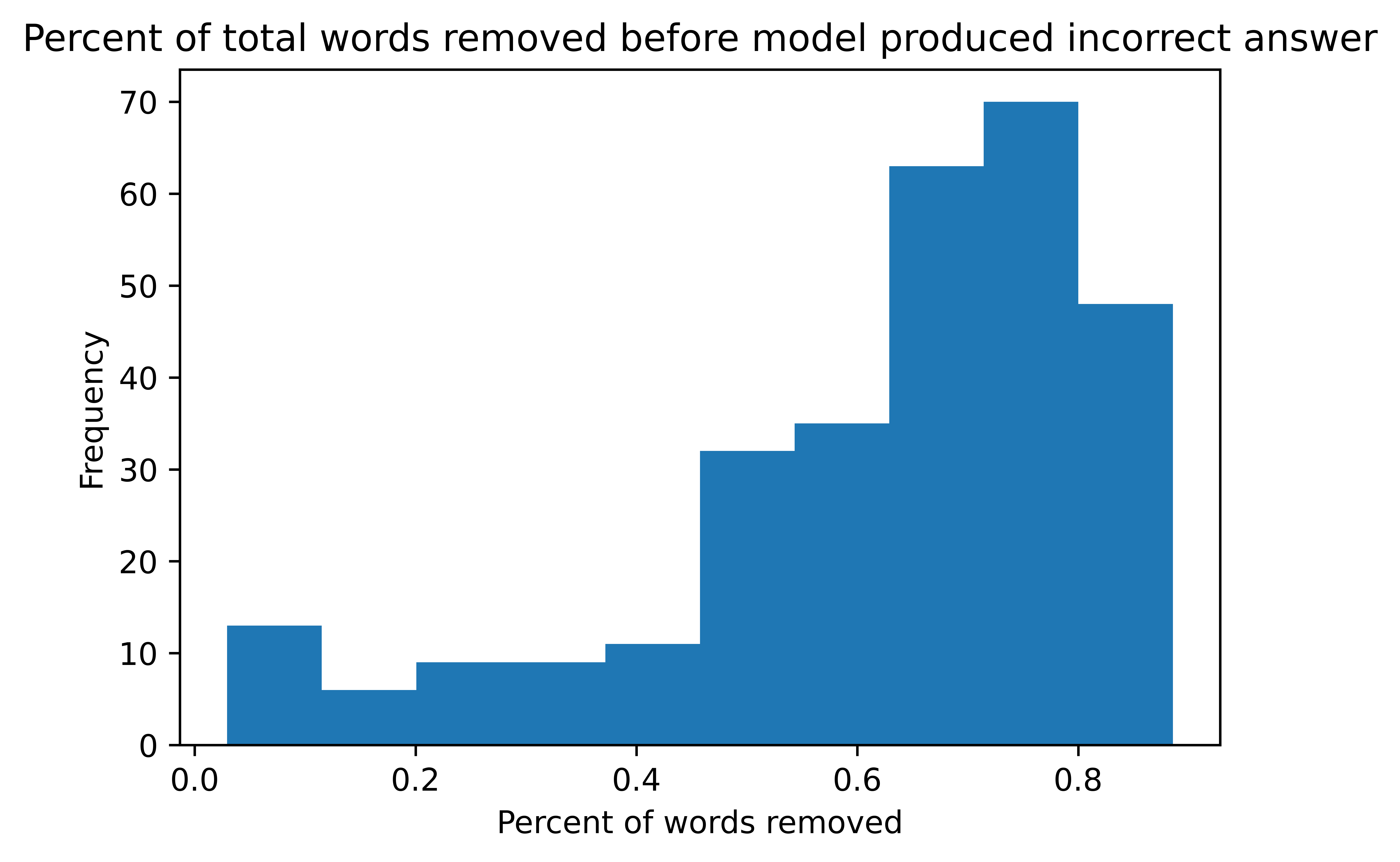}
  \caption{A histogram of the percentage of words in a given MWP were removed before the model produced an incorrect solution to the problem. This histogram does not include MWPs that the model gave an incorrect prediction with the original text.}
  \label{fig:input_red_hist}
\end{figure}

In this experiment, we examine the diversity, or lack thereof, of words in the MaWPS' dataset's vocabulary. Our work is intended to reveal possible trigger words that may frequently appear in some types of problems but not others.

\subsection{Methods}

We looked at the word frequency of words in the first cross-validation fold of the MaWPS dataset, both the training and testing datasets. The problem texts were first set to all lowercase letters, then stemmed and lemmatized in order to count all occurrences of the words.

We counted the number of MWPs that each word appeared in rather than the total number of appearances of each word. We found the top 50 words, by number of MWPs the word appeared in, for every operation type (+, -, *, /, multiple), then filtered out any words that appeared in every list. In this way we can see which words are uniquely frequent in specific operations, and are not just frequent in the corpus overall. 

\subsection{Results}

The results are shown in Table \ref{tab:top_words}. We can see that these words often appear in 10-20\% of all problems of a given type, though the majority of the words do not appear to have any correlation to the type of operation that they most often appear in. 

\subsection{Discussion}

None of the most popular words appeared to be relevant to the category of problem that they most frequently appeared in. The fact that these words are appearing so frequently indicates a low lexical diversity in the MaWPS dataset, which may encourage the model to rely on the occurrence of these words to classify problems into different operations.

\section{Experiment 3: Input Reduction}

We used input reduction to uncover how many words can be removed from an MWP before the model will produce an incorrect answer. If very few words remain and have little to do with the correct equation, it suggests that the model is not performing much semantic reasoning between quantities in order to find the correct equation.

\subsection{Methods}

Our approach is based on the work of \citeauthor{feng_pathologies_2018} (\citeyear{feng_pathologies_2018}), but does not follow their exact methodology. We implemented confidence scores using the posterior probability of each label, summed those probabilities and divided by the number of outputs, since we were using an RNN model. For the input reduction process, we iteratively removed the word which reduced the model's confidence score the least.

We used only the RNN Seq2seq model created from the first CV fold of MaWPS for our input reduction predictions.

\subsection{Results}

A histogram of the percentage of words removed when the model gave an incorrect prediction is shown in Figure \ref{fig:input_red_hist}. The histogram does not include MWPs that the model got wrong with the original text. The mean percentage of words removed is 62.3\%, while the median is 68.1\%. This means that the model produces the correct prediction with less than 68.1\% of the words for half of the problems it is able to solve. 

An example of the input reduction process is shown in Figure \ref{tab:input_red_ex}. In this example, 22 words are removed before the model produces an incorrect equation. The most reduced input to receive a correct equation is "his number0 each his number1 many," which arguably contains little to no information about what the correct equation is, and yet the model still solves the problem with high confidence (99\%).

\begin{table*}
\centering
\begin{tabular}{>{\centering\arraybackslash}p{2cm}crlp{7.5cm}}
\midrule\midrule
    &   Score &   Model Confidence & Removed Word   & Question                                                                                                                              \\
\hline
  0 &      Correct &           0.999997 & nan            & Emily collects number0 cards . Emily 's father gives Emily number1 more . Bruce has number2 apples . How many cards \textcolor{red}{does} Emily have ? \\
  1 &      Correct &           0.999999 & does           & Emily collects number0 cards . Emily 's father gives Emily number1 more . Bruce \textcolor{red}{has} number2 apples . How many cards Emily have ?      \\
  2 &      Correct &           0.999999 & has            & Emily collects number0 cards . Emily 's father gives Emily number1 more . Bruce number2 apples . \textcolor{red}{How} many cards Emily have ?          \\
  3 &      Correct &           0.999999 & how            & Emily \textcolor{red}{collects} number0 cards . Emily 's father gives Emily number1 more . Bruce number2 apples . many cards Emily have ?              \\
  4 &      Correct &           0.999998 & collects       & Emily number0 cards . Emily \textcolor{red}{'s} father gives Emily number1 more . Bruce number2 apples . many cards Emily have ?                       \\
  5 &      Correct &           0.999998 & 's             & \textcolor{red}{Emily} number0 cards . \textcolor{red}{Emily} father gives \textcolor{red}{Emily} number1 more . Bruce number2 apples . many cards \textcolor{red}{Emily} have ?                          \\
  6 &      Correct &           0.999998 & emily          & number0 cards . father \textcolor{red}{gives} number1 more . Bruce number2 apples . many cards have ?                                                  \\
  7 &      Correct &           0.999997 & gives          & number0 cards . father number1 \textcolor{red}{more} . Bruce number2 apples . many cards have ?                                                        \\
  8 &      Correct &           0.999995 & more           & number0 cards . \textcolor{red}{father} number1 . Bruce number2 apples . many cards have ?                                                             \\
  9 &      Correct &           0.999995 & father         & number0 cards . number1 . Bruce number2 apples . many cards \textcolor{red}{have} ?                                                                    \\
 10 &      Correct &           0.999994 & have           & number0 cards . number1 . Bruce number2 apples . \textcolor{red}{many} cards ?                                                                         \\
 11 &      Correct &           0.999993 & many           & number0 cards . number1 . Bruce number2 apples . cards \textcolor{red}{?}                                                                              \\
 12 &      Correct &           0.999968 & ?              & number0 \textcolor{red}{cards} . number1 . Bruce number2 apples . \textcolor{red}{cards}                                                                                \\
 13 &      Correct &           0.995461 & cards          & number0 . number1 . \textcolor{red}{Bruce} number2 apples .                                                                                            \\
 14 &       Incorrect &           0.944366 & bruce          & number0 . number1 . number2 apples .     \\
\midrule
\midrule
\end{tabular}
\caption{An example of the input reduction process.}
\label{tab:input_red_ex}
\end{table*}

\subsection{Discussion}

The results of the input reduction experiment show that in most cases more than half of the total words can be removed from the MWP before the model produces an incorrect answer. With over half of the words removed, these problems are nonsensical to humans, as in Table \ref{tab:input_red_ex}. This indicates that the model is not truly performing reasoning about the sequence of events explained in the problem, since it can still produce a correct equation with over half of the information removed from the input.

\section{Future Work}

We would like to implement the gradient-based method used by \citeauthor{feng_pathologies_2018} (\citeyear{feng_pathologies_2018}) in order to obtain a more objective idea of how much removing a given word affects the model. The current confidence score approach produces very high confidence on almost every input, even when it is wrong, which reduces the credibility of our input reduction results.

We would also like to implement the high-entropy output fine-tuning suggested by \citeauthor{feng_pathologies_2018} (\citeyear{feng_pathologies_2018}) to possibly improve the interpretability and accuracy of the RNN Seq2seq MWP solver.

Another possible avenue of word would be to increase the lexical diversity of MaWPS by writing code to change words to synonyms before the MWPs are fed into the model for training. This way, the model would not be able to rely on the high frequency of certain words to make its predictions.

\balance

\section{Conclusion}

The results of Experiment 1, parts of speech removal, indicated a small reliance on some parts of speech, especially nouns and verbs. The AsDIV-A model was also shown to be more reliant on specific parts of speech than MaWPS, perhaps indicating some overfitting to those words. Experiment 2, word frequency in MaWPS, shows that the lexical diversity of MaWPS is low. Experiment 3, input reduction, shows that well over half of the words in a given MWP can be removed before the model gives an incorrect prediction. This shows that the model is not using all of the information in the question to make its prediction, and may be relying on occurrences of some of the words from Experiment 2, or some other superficial patterns, to make its predictions.

\begin{acks}

The work reported in this paper is supported by the National Science Foundation under Grant No. 2050919. Any opinions, findings, and conclusions or recommendations expressed in this work are those of the author(s) and do not necessarily reflect the views of the National Science Foundation.

\end{acks}

\bibliographystyle{ACM-Reference-Format.bst}
\bibliography{explainability}

\end{document}